\documentclass{ifacconf}  

\usepackage{cite}
\usepackage{algorithm}
\usepackage[noend]{algpseudocode}
\usepackage{textcomp}
\usepackage{xcolor}
\usepackage{comment}
\usepackage{float}

\usepackage{graphicx} 
\usepackage{amsmath} 
\usepackage{amssymb}
\usepackage{comment}
\usepackage{multirow}
\usepackage{color}
\usepackage{natbib}        

\begin{document}
\begin{frontmatter}

\title{\LARGE \bf
Split-Boost Neural Networks
}

\thanks{Corresponding author: {\tt simone.formentin@polimi.it}.}
\thanks[footnoteinfo]{This project was partially supported by the Italian Ministry of University and Research under the PRIN'17 project \textquotedblleft Data-driven learning of constrained control systems", contract no. 2017J89ARP.}

\author[First]{Raffaele G. Cestari} 
\author[Second]{Gabriele Maroni}
\author[Second]{Loris Cannelli}
\author[Second]{Dario Piga}
\author[First]{Simone Formentin}

\address[First]{Department of Electronics, Information, and Bioengineering, Politecnico di Milano, Milano, Italy.}
\address[Second]{SUPSI-DTI-IDSIA, Dalle Molle Institute for Artificial Intelligence, Lugano, Switzerland.}

\maketitle
\thispagestyle{empty}
\pagestyle{empty}

\begin{abstract}
The calibration and training of a neural network is a complex and time-consuming procedure that requires significant computational resources to achieve satisfactory results. Key obstacles are a large number of hyperparameters to select and the onset of overfitting in the face of a small amount of data. In this framework, we propose an innovative training strategy for feed-forward architectures - called \textit{split-boost} - that improves performance and automatically includes a regularizing behaviour without modeling it explicitly. Such a novel approach ultimately allows us to avoid explicitly modeling the regularization term, decreasing the total number of hyperparameters and speeding up the tuning phase. The proposed strategy is tested on a real-world (anonymized) dataset within a benchmark medical insurance design problem.
\end{abstract}

\begin{keyword}
Artificial intelligence, deep learning, machine learning, neural networks, hyperparameter tuning, regularization
\end{keyword}
\end{frontmatter}
\section{INTRODUCTION}
Training a neural network is a complex task for several reasons. The high dimensionality of the search space in which the network parameters live (number of layers, number of neurons per layer, learning rate, choice of activation function, batch size, and regularization factor) makes the problem difficult to optimize. The optimization problem for network parameters update is non-convex and must be solved with gradient descent strategies that might suffer from the presence of local minima and saddle points that slow down the convergence to global optima. Very often it happens that the choice of a hyperparameter affects the value that the others must assume and the change of one of them can lead to unexpected changes in the behavior of the network. In addition, since training a neural network is computationally expensive (even if the performance in terms of Graphic Processing Unit (GPU) computing power is increasingly reducing training times \cite{Gupta}, \cite{Sutskever}) hyperparameter space exploration might be a slow and complex task. Lastly, although there are guidelines for selecting hyperparameters, in general, there is not one specific solution to the tuning problem and the quality of the selection might depend on the specific case, considering the dataset and network architecture. 

Given these major challenges, we propose an innovative feed-forward neural network training strategy. The goal is twofold:
\begin{itemize}
    \item removing the calibration step of the regularization term, otherwise indispensable to avoid the onset of overfitting, facilitating the training procedure of the network;
    \item include regularization \textit{automatically} in the training process exploiting different training data. The goal is to achieve a regularization effect without directly adding a corresponding term to the cost. 
\end{itemize}

The way these goals are pursued gives rise to the name we came up with for the new training  strategy proposed in this contribution, namely \textit{split-boost neural network}. 

The first phase involves dividing the training set into $k$ equal parts. In this, we were inspired by the renowned idea of k-fold cross validation (see \cite{Geisser}, \cite{Kohavi}, \cite{Varma}, \cite{Arlot}). From this procedure originates the name \textit{split}. In this work, $k$ is chosen  equal to $2$, but other values can be selected without any loss of generality. 
We consider a network structure with  two fully-connected layers. The idea is to update the parameters of the first layer with gradient descent, explicitly calculating the terms that yield the gradient as a function of optimal values of the weights of the second layer. These weights are identified separately, using each time one portion of the training set. This is what leads the name \textit{boost}. This procedure \textit{boosts} training performances and includes automatic regularization in the network to avoid overfitting. The weights of the second layer, calculated independently on the two portions, are then \textit{averaged} to identify the optimal value to use in prediction. This methodology finds similarities with the panorama of Extreme Learning Machines (ELMs), see \cite{Huang}, in which the optimal values of the parameters of the second layer given those of the first are explicitly obtained. However, our work differs from traditional ELMs, where the parameters of the first layer are fixed a priori (and not updated) after initial randomization, see \cite{Huang2}. Additionally, backpropagation in ELMs is removed. Instead, we still perform backpropagation to train the first layer's parameters. Likewise, a one-shot optimization is performed (ridge regression) with the substantial difference in how the data are divided to parallelize the computation on 2 different batches to retrieve the parameters of the second layer explicitly. This strategy is expected to reduce the occurrence of overfitting without resorting to a regularization term.

The paper is organized as follows. In Section \ref{pbstatement} a review of the state of the art, the main challenges, limitations, and the benchmark network structure are presented. In Section \ref{mathForm} the mathematical formulation of the training strategy of the \textit{split-boost} neural network is presented: in the first part of the section the mathematical notation is introduced, then in Section \ref{optimiz} the \textit{split-boost} optimization problem for optimal weights computation is defined, and finally in Section \ref{details} the details on training procedure, the policy for best epoch retrieval and the learning rate switching strategy are described. In Section \ref{results} a numerical comparison between the traditional feed-forward neural network training and the split-boost training is presented. 
The paper is ended by some concluding remarks.

\section{PROBLEM STATEMENT}
\label{pbstatement}
The goal of this work is to propose an alternative training strategy for a classic feed-forward neural network. Without modification of the network structure, we want to show that using the same amount of data in a different way allows us to improve training performances and achieve implicit regularization, namely to obtain a regularization effect without modeling it explicitly. This allows us to simplify the tuning procedure of the network, reducing the number of hyperparameters to be calibrated.

In this section, the basic structure of the architecture of a feed-forward neural network is introduced without going into details as this morphology is widely studied in the literature, see \cite{Rosenblatt}, \cite{Bishop}, \cite{Nair}, \cite{Goodfellow}. A feed-forward neural network is a deep learning model which, based on the interaction of several processing units, called neurons, introducing non-linearities on the inputs, can perform various tasks ranging from classification, regression and prediction of time series. The parameters that build this model are represented by the weight matrices that correspond to the different layers of neurons building the architecture. These weights are updated through  epochs (e.g., several passes through the dataset) via gradient descent (or one of its variants) and the help of the well-known backpropagation algorithm \cite{Rumelhart}, for the calculation of the gradient of the errors with respect to the weights of the network.

Neural networks are models with a high descriptive capacity but are characterized by the phenomenon of overfitting, which can cause negative repercussions on the generalization capacity.
Overfitting is one of the curses of general statistical learning. It often discourages users of artificial intelligence as the lack of a sufficient amount of data makes the architectures prone to its onset. In the literature, there are several documents relating to the description of the problem (see \cite{hinton2}, \cite{Krizhevsky}, \cite{Keskar}, \cite{Bengio}).  
In the case of feed-forward neural networks, there are several strategies that can counter overfitting such as dropout, early stopping, regularization, data augmentation, batch normalization (see, e.g., \cite{Srivastava}, \cite{Ioffe}, \cite{Hinton},  \cite{Loshchilov}).  
The goal of these methodologies is to prevent the neural network from overly relying (e.g. fitting the noise) on the data used in training. To do this, the idea is to reduce the number of network hyperparameters (or limit their norm) or artificially increase the available data. In this study, we assume \textit{regularization} as a benchmark of anti-overfitting methodology. This technique consists in adding into the cost function that must be minimized (or equivalently the reward that must be maximized) a term that depends on the norm of the weights of the different layers of the neural network. During the training step, the neural network is forced to keep the norm of these weights limited to prevent an excessive increase in the cost term.

\begin{figure}[h!]
\centering
\includegraphics[width=9cm]{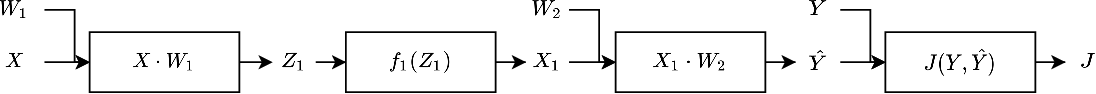}
\caption{Neural network architecture}
\label{fig:NN_architecture}
\end{figure}

In Figure \ref{fig:NN_architecture} 
a sketch of the neural network architecture in matrix notation used in this work is shown. 
The considered structure consists of 2 layers (hidden and output) and is designed to solve a regression problem.
Each input is processed by each layer (and by each neuron per layer) through the non-linear activation function $f_1$ (rectified linear unit, RELU). $X \in \mathbb{R}^{N \times D}$ is the input data matrix, where $N$ is the number of samples and $D$ is the number of features. $W_1 \in \mathbb{R}^{D \times H}$ is the weight matrix of the first (hidden) layer, where $H$ is the number of neurons. $W_1^b \in \mathbb{R}^{H \times 1}$ is the bias vector of the first (hidden) layer. $Z_1 \in \mathbb{R}^{N \times H}$ are the pre-activations of the first layer. $X_1 \in \mathbb{R}^{N \times H}$ are the activations of the first layer. $W_2 \in \mathbb{R}^{H \times 1}$ is the weight matrix of the second (output) layer. $W_2^b \in \mathbb{R}$ is the bias of the output layer. $Y,\hat{Y} \in \mathbb{R}^{N \times 1}$ are respectively the matrix of the targets and the matrix of the prediction, and finally, $J$ is the loss function. Layer biases $W_1^b$ and $W_2^b$ follow the same training procedure of the corresponding layer weights. For sake of compactness of notation, they are omitted.

The optimization problem that must be solved to find the optimal weights of a feed-forward neural network with 2 fully connected layers, following the traditional training procedure known in literature has the structure of the following unconstrained minimization problem:
\begin{equation}
\label{feedforward}
    \min_{W_1,W_2}{\frac{1}{2N_t}||Y_t -f_1(X_t\cdot W_1)\cdot W_2||_2^2 + \frac{\lambda}{2}\sum_{i=1}^2||W_i||_2^2}
\end{equation}
where the subscript $t$ refers to the training set and $\lambda$ is a hyperparameter that controls the intensity of the regularization. In the next section, we present an alternative, novel, way to write down the same optimization problem for training a 2-layer fully connected neural network in a way that {allows us to exploit at best the training information introducing an implicit regularizing effect and simultaneously reducing the number of parameters to be tuned. Indeed, within this framework, the regularization factor is not needed anymore.}

\section{MATHEMATICAL FORMULATION OF SPLIT-BOOST NETWORKS}
\label{mathForm}
This section illustrates this alternative training strategy of the proposed Split-Boost Neural Network. The updating of the weight parameters takes place separately for the $2$ layers (hidden and output) that characterize the network architecture. The idea is to formulate   the training procedure as a bilevel optimization problem, whose outer optimization variables are the weights of the hidden layer $W_1$, and the inner optimization variables are the weights of the output layer $W_2$. For a regression problem, the optimal values of the parameters of the output layer $W_2$ can be obtained in closed form by solving two least square problems. These least square problem solutions represent the constraints of the first optimization problem. 

The algorithm involves first a  \textit{splitting step}, in which the training set is divided into two sub-sets (a reasonable choice is to divide it equally, we do not claim that this choice is the only possible one, however, it guarantees that both optimization problems see the same amount of data, avoiding the unbalancing towards one of the two partitions.). Both subsets are then used to solve, separately, two least squares problems. Once the least squares problems are solved (e.g. the optimal values for $W_2$ with respect to the two sub-sets are found), the whole training set is used to update the values of $W_1$. 

From here, \textit{the boosting idea}. The optimal values obtained with the first sub-set are used to generate the prediction for the data belonging to the second sub-set and vice-versa. Our goal is to show that this methodology can effectively replace the regularization term in a traditional feed-forward neural network, overcoming its performance. This step represents one of the main differences compared to traditional network training: dividing the training set into 2 batches, used to calibrate the parameters of the second layer independently, allows us to improve the information content extraction.
In Table \ref{tabella1} the symbols and their description is summarized. 
\begin{table}[h]
\centering
\caption{Notation.}
\begin{tabular}{|l|l|}
\hline
\textbf{\textbf{Symbol}} & \textbf{Description} \\
\hline
$N$ & Number of samples \\
$t,v,ts$ & Training, validation and test sets \\
$a$ & Training set partition ``A'' \\
$b$ & Training set partition ``B'' \\
$D$   & Number of input features \\
$X \in \mathbb{R}^{N \times D}$ & Input data \\
$Y,\hat{Y} \in \mathbb{R}^{N \times 1}$   & Output data, prediction \\
$Z \in \mathbb{R}^{N \times H}$ & Pre-activations \\
$X_1 \in \mathbb{R}^{N \times H}$ & Activations\\
$H$ & Number of neurons per layer \\
$W_1 \in \mathbb{R}^{D \times H}$ & Weights of hidden layer\\
$W_2 \in \mathbb{R}^{H \times 1}$ & Weights of output layer\\
$f_1:\mathbb{R}^{N \times H} \rightarrow \mathbb{R}^{N \times H}$ & Activation function \\
$J$ & Cost function\\
$\lambda$ & Regularization parameter \\
$\gamma$ & Learning rate \\
\hline
\end{tabular}
\label{tabella1}
\end{table}

\subsection{Optimization problem, forward and backward propagations.}
\label{optimiz}
The heart of the training algorithm is represented by the optimization problem shown in Equation \eqref{optimization}. 
Since this cannot be solved in closed form, it is solved iteratively through a descending gradient problem. The value of the $W_1$ parameters is therefore updated as the epochs pass.
\begin{subequations}
\label{optimization}
\begin{align}
\label{totalcost}
\min_{W_1} \,\,&J=\frac{1}{2N_b}\|Y_b-f_1(X_b\cdot W_1)\cdot W_{2a}^*(W_1)\|_2^2 \nonumber \\
&+ \frac{1}{2N_a}\|Y_a-f_1(X_a\cdot W_1)\cdot W_{2b}^*(W_1)\|_2^2\\
W_{2a}^*(W_1) &= \operatorname*{argmin}_{W_2}\frac{1}{2N_a}\|Y_a-f_1(X_a\cdot W_1)\cdot W_2\|_2^2\\
W_{2b}^*(W_1) &= \operatorname*{argmin}_{W_2}\frac{1}{2N_b}\|Y_b-f_1(X_b\cdot W_1)\cdot W_2\|_2^2
\end{align}
\end{subequations}

{In what follows, the subscripts $k,j$ are used to refer both to training subsets A and B. Since the description of the equations referring to the two training sets mirror each other, the notation will follow as $\forall k,j \in [a,b], k \neq j$. All the procedure is repeated for both admissible values of $k$ and $j$, training is \textit{splitted}. The subscript notation is summarized in Table \ref{tabellaSubscript}.} 

\begin{table}[h!]
\centering
\begin{tabular}{|l|l|l|}
\hline
\textbf{Subscript} & \textbf{Set A} & \textbf{Set B}\\
\hline
$k$ & $a$ & $b$ \\
$j$ & $b$ & $a$ \\
\hline
\end{tabular}
\caption{Training sets notation.}
\label{tabellaSubscript}
\end{table}

In the following equations,  the forward propagation is described. We do not go into details as this step is well-known in literature (see \cite{Rumelhart}, \cite{Werbos}, \cite{Fahlman}, \cite{Bartlett}, \cite{LeCun}).
\begin{subequations}
\label{forward}
\begin{align}
Z_{1k} &=X_k \cdot W_1\\
X_{1k} &=f_1(Z_{1k}) \\
\hat{Y}_k &= X_{1k} \cdot W_{2j}\\
J_{W_{2k}} &= \frac{1}{2N_k}\|Y_k-\hat{Y}_k\|_2^2
\end{align}
\end{subequations}
We can compute explicitly the optimal values for the output layer parameters $W_2$ solving a least square problem:  
\begin{equation}
\label{optimalW2}
        W_{2k}^*=(X^T_{1k} \cdot X_{1k}  )^{-1} \cdot X^T_{1k}Y_k. 
\end{equation}

Solving explicitly Equation \eqref{optimalW2} we derive the optimal values for the output layer parameters computed with the forward pass in the two separate training sub-sets. Notice that $W_{2k}^*$ is function of $W_1$ through the dependency on $X_{1k}$. For sake of brevity in the following derivation, we write $W_{2k}^*$ in place of $W_{2k}^*(W_1)$.
For sake of compactness, the Jacobian expression is derived in scalar form. Nonetheless, it must be interpreted as the matrix of first-order partial derivatives of a vector-valued function with respect to its input variables.
Since $W^*_{2k}$ is an optimum computed explicitly, it is true that:
\begin{align} \label{eqn:partial_w2}
\frac{\partial J_k(W_1, W_{2k})}{\partial W_{2k}} \bigg\rvert_{W_{2k} = W_{2k}^*} = \frac{\partial J_k(W_1, W_{2k}^*)}{\partial W_{2k}^*} = 0
\end{align}
Differentiating both sides with respect to $W_1$ and using the chain rule according to the influence diagram in Figure \ref{fig:influence_diagram}:
\begin{figure}[h!]
\centering
\includegraphics[width=7cm]{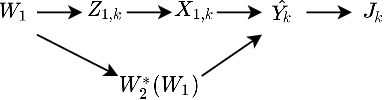}
\caption[Influence diagram.]{Influence diagram describing the interaction between network layer parameters.}
\label{fig:influence_diagram}
\end{figure}
\begin{gather} 
\begin{subequations}
	\label{eqn:partial_w2_w1}
\begin{align}
&\frac{\partial}{\partial W_1}\left(\frac{\partial J_k(W_1, W_{2k}^*)}{\partial W_{2k}^*} \right) = 0  \nonumber \\
&\frac{\partial}{\partial W_1}\left(\frac{\partial J_k(W_1, W_{2k}^*)}{\partial W_{2k}^*} \right) + \nonumber \\
&+ \frac{\partial}{\partial W_{2k}^*}\left(\frac{\partial J_k(W_1, W_{2k}^*)}{\partial W_{2k}^*} \right) \cdot \frac{\partial W_{2k}^*}{\partial W_1} = 0  \nonumber \\
&\frac{\partial^2 J_k(W_1, W_{2k}^*)}{\partial W_1 \partial W_{2k}^*} + \frac{\partial^2 J_k(W_1, W_{2k}^*)}{\partial W_{2k}^* \partial {W_{2k}^*}^\text{T}} \cdot \frac{\partial W_{2k}^*}{\partial W_1} = 0 \nonumber
\end{align}
\end{subequations}
\end{gather}
Solving for the Jacobian $\frac{\partial W_{2k}^*}{\partial W_1}$:
\begin{align} \label{jacobiano1}
\frac{\partial W_{2k}^*}{\partial W_1} = - \left(\frac{\partial^2 J_k(W_1, W_{2k}^*)}{\partial W_{2k}^* \partial {W_{2k}^*}^\text{T}} \right)^{-1} \cdot \frac{\partial^2 J_k(W_1, W_{2k}^*)}{\partial W_1 \partial W_{2k}^*}
\end{align}
And then with notation abuse (we use the approximation symbol to treat the Jacobian as in the scalar case even though, as commented earlier, it must be interpreted as the matrix of first-order partial derivatives): 
\begin{equation}
\label{jacobiano2}
\mathbf{J}_{W_1}W^*_{2k}\approx\frac{\partial W_{2k}^*}{\partial W_1}
\end{equation}
The Jacobian $\mathbf{J}_{W_1}W^*_{2k}$ described in Equations \eqref{jacobiano1} and \eqref{jacobiano2} is a fundamental ingredient in the evaluation of the gradient of the cost function $J$, to update $W_1$ values.
To complete the training procedure, the backward propagation step must be performed. Similarly to forward propagation, backward propagation also takes place by working simultaneously on both sub-sets of the training set. This allows us to derive the expressions of the gradients necessary for the final calculation of the gradient of $J$ with respect to $W1$: 
\begin{subequations}
\label{backprop}
    \begin{align}
        \nabla_{\hat{Y}_k}J_k&=\frac{1}{N_k}(\hat{Y}_k-Y_k)\\
        \nabla_{X_{1k}}J_k&=\nabla_{\hat{Y}_k}J_k \cdot W^{*^T}_{2j}\\
        \nabla_{W^*_{2j}}J_k&=X^T_{1k} \cdot \nabla_{\hat{Y}_k}J_k\\
        \nabla_{Z_{1k}}J_k&=\nabla_{X_{1k}}J_k \circ f_1'(Z_{1k})
    \end{align}
\end{subequations}
For a deeper understanding of how backpropagation works, refer to \cite{Hecht-Nielsen}, \cite{Bottou}, \cite{Glorot}. 

Merging together the results obtained by solving Equations \eqref{forward}, \eqref{optimalW2} and \eqref{backprop} we derive the expression of the gradient of the cost function \eqref{totalcost} with respect to the weights of the first layer, $W_1$, described in the following equation: 
\begin{subequations}
\label{update}
\begin{align}
\label{totalGradient}
    \nabla_{W_1}J=&X_b^T\nabla_{Z_{1b}}J_b + \mathbf{J}_{W_1}W^{*^T}_{2a}\nabla_{W^*_{2a}}J_b+\nonumber \\
    +&X_a^T\nabla_{Z_{1a}}J_a + \mathbf{J}_{W_1}W^{*^T}_{2b}\nabla_{W^*_{2b}}J_a\\
    \label{w1up}
    W_1 &= W_1 - \gamma\nabla_{W_1}J\\
    \label{w2up}
    W_2 &= \frac{W^*_{2a}+W^*_{2b}}{2}
\end{align}
\end{subequations}
Thanks to Equations \eqref{update} and \eqref{optimalW2} we can update network parameters between two consecutive epochs.
Notice the difference: $W_1$ is updated through gradient descent in Equation \eqref{w1up} while $W_2$ is obtained as the average of the optimal $W_2^*$ computed explicitly by looking at the two training sub-sets separately in Equation \eqref{w2up}. Notice that, in general, this closed-form solution for $W_2$ might not be practicable (e.g. in the case of classification problems, the concatenation of non-linearities through the different layers of the network would make it unattainable to formulate a least squares problem, as in this case, efficiently solved analytically). In that case, a similar expression to \eqref{totalGradient} must be derived and update $W_2$ with gradient descent. However, for this setting, it improves the computational time. 
Equations \eqref{update} represent the heart of the algorithm, defining the \textit{boosting} step. The mixing of the two sub-sets of the training set embedded in the gradient expression and in optimal values for $W_2$ can improve training performance achieving the same training cost in a lower number of epochs and at the same time avoiding overfitting without the aid of regularization terms. This is critical for reducing the number of hyperparameters.
With the updated weights obtained in Equations \eqref{update}, the prediction is computed as follows: 
\begin{equation}
    \hat{Y}=f_1(X\cdot W_1)W_2
\end{equation}
\subsection{Details on training procedure}
\label{details}
Early stopping procedure is used.
The stop condition for training is obtained by monitoring the status of the validation cost, as follows:
\begin{equation}
    |J_v(k) -J_v(k-1)|<\epsilon
\end{equation}
where $k$ is the epoch index and $\epsilon$) is the stop threshold. If the variation of the validation cost between two consecutive epochs is less than $\epsilon$, training stops and the optimal number of epochs is retrieved. After the computation of the optimal number of epochs, the network is re-trained using as new training set the sum of the original training set, and the validation set. Performances are evaluated considering the test set.

The split-boost strategy has a learning rate varying with the number of epochs. 
This choice avoids oscillations in training cost (an excessively large learning rate leads to fluctuations and consequent degradation of performance). In the example reported in the next section, the following switch condition is adopted:
\begin{equation}
    \begin{cases}
    \gamma = \gamma^* \,\,\, if \,\,J_t(k) - J_t(k-1) > 0\\
    \gamma = \frac{\gamma^*}{10} \,\,\, otherwise
    \end{cases}
\end{equation}
where $\gamma^*$ is the best learning rate, chosen after a sensitivity analysis carried on the validation cost.

\section{AN EXPERIMENTAL CASE STUDY}
\label{results}
In this section, we show the comparison between a traditional feed-forward neural network and the split-boost neural network applied to a real-world regression problem. 
The code is written in Python 3.7.13. The Python library used to develop the neural networks is PyTorch \cite{pytorch}. Simulations run on an Intel Core i7-8750H with 6 cores, at 2.20 GHz (maximum single core frequency: 4.10 GHz), with 16 GB RAM.
\subsection{Dataset description}
\label{dataset}
The case-study is the medical insurance forecast of patients living in the U.S., given a set of clinical features. Data are open-source and offered in \cite{book}. In Table \ref{dataFeatures} data features and targets are summarized. The goal is to predict the medical insurance charge ($\$$) given a set of $D=6$ features: age, sex, BMI, number of children, smoking condition, and region of residence. 

\begin{table}[h]
\label{dataFeatures}
\centering
\begin{tabular}{|l|l|}
\hline
\textbf{\textbf{Features}} & \textbf{Description}   \\
\hline
Age & Age of primary beneficiary   \\
Sex & Insurance contractor gender  \\
BMI & Body mass index   \\
Children & Number of children \\
Smoker   & Smoker condition \\
Region & Residential area in the U.S.\\
\hline
\textbf{Target} & \textbf{Description} \\ 
\hline
Charge & Medical insurance bill [\$]  \\
\hline
\end{tabular}
\caption{Medical Insurance Dataset: features and target.}
\end{table}

\begin{figure}[h!]
\centering
\includegraphics[width=9cm]{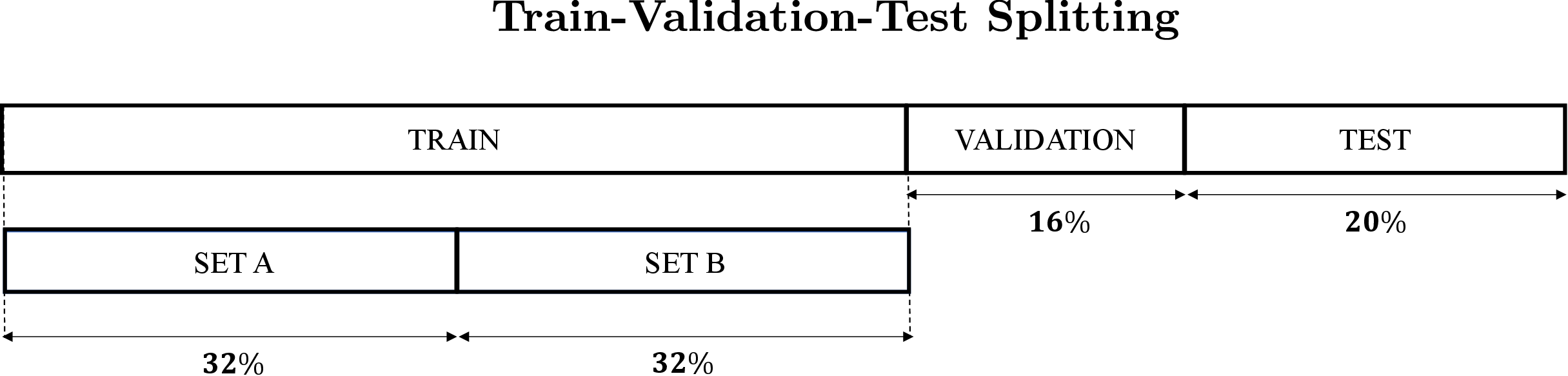}
\caption{Train-Validation-Test splitting. The training set is divided into two sub-sets A and B. }
\label{fig:splitting}
\end{figure}

The number of people in the study is $N=1338$. Dataset is split into training, validation and test sets according to the proportions shown in Figure \ref{fig:splitting}. Test set is of $20\%$ of the total. The validation set is $16\%$. The training set is $64\%$. In the case of the split-boost neural network, the training set is further divided into two halves, $32\%$ each.

\subsection{Networks Hyperparameters}
\label{hyperparameters}
\begin{figure}[h!]
\centering
\includegraphics[width=1\columnwidth]{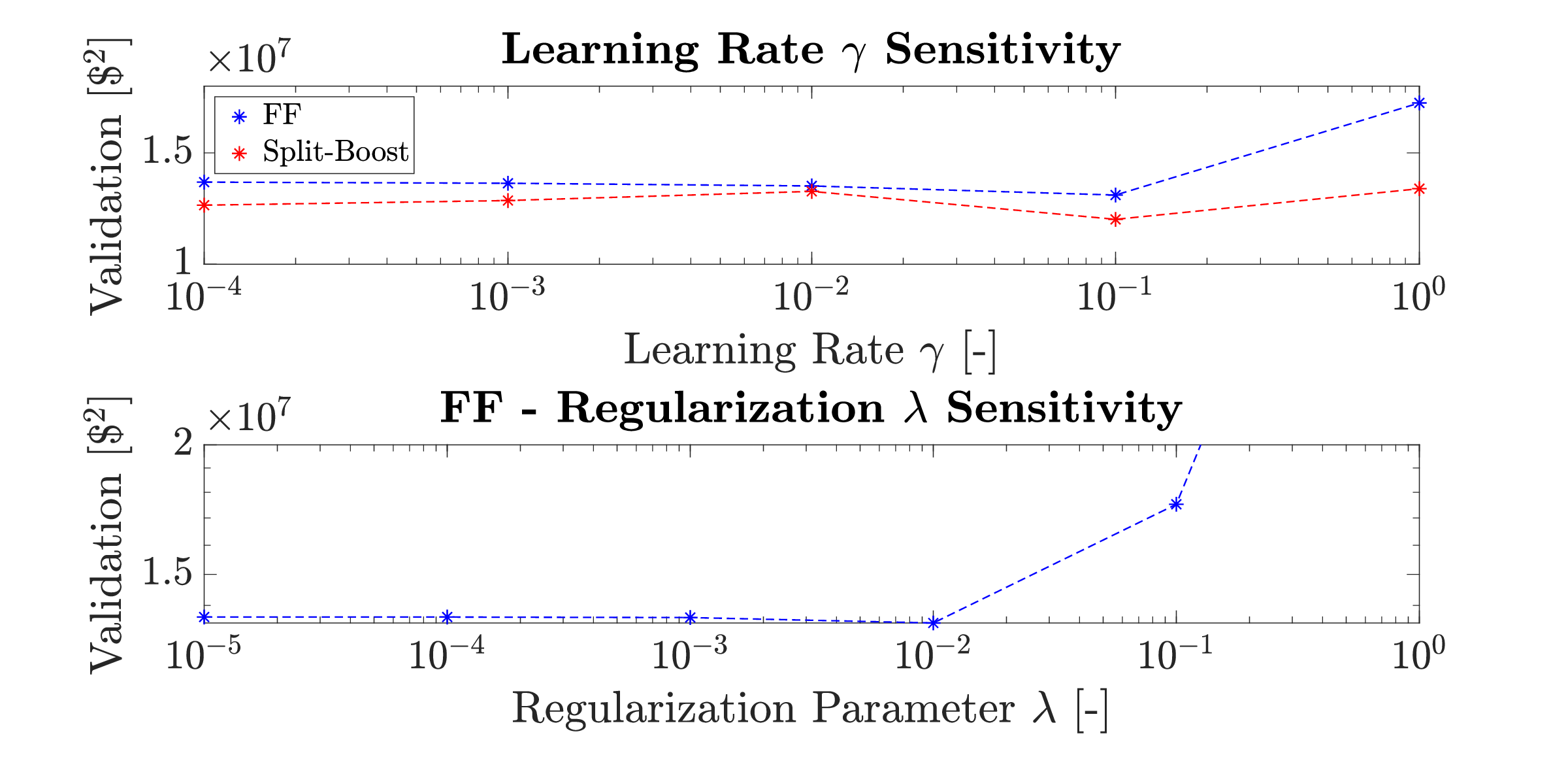}
\caption{Hyperparameter tuning:  sensitivity analysis with respect to the learning rate $\gamma$ (upper panel);   sensitivity analysis with respect to the  regularization parameter $\lambda$ (lower plot).} 
\label{fig:tuning}
\end{figure}

In this section, we show the hyperparameter tuning procedure. Sensitivity analysis of the learning rate $\gamma$ and the regularization factor $\lambda$ is performed. In the case of the split-boost neural network, there is the advantage of not having a regularization term which, instead, is replaced by the boosting procedure. 
Sensitivity analysis with respect to $\gamma$ is described in the upper plot of  Figure \ref{fig:tuning}. The validation cost $J_{v}= \frac{1}{2N_v}(Y_v-\hat{Y}_v)^2$ associated to different choices of $\gamma$ is shown. The best choice for both networks is $\gamma=0.1$. In the lower plot of Figure \ref{fig:tuning} the sensitivity analysis with respect to  the regularization hyper-parameter $\lambda$  is shown (only for the feed-forward neural network), evaluating the validation cost. The goal is to compare the best possible feed-forward regularized neural network with the split-boost network. The best choice is $\lambda = 0.01$.
In Table \ref{tabellaHyper} the network parameters are summarized. Network architecture is the same. Same number of layers ($2$), neurons per layer ($H=10$) and activation function (RELU). Split-boost strategy does not need any regularization parameter. 
\begin{table}[h!]
\centering
\begin{tabular}{|l|l|l|}
\hline
\textbf{Hyperparameter} & \textbf{FF} & \textbf{Split-Boost}\\
\hline
$L$ (Layers) & $2$ & $2$ \\
$H$ & $10$ & $10$ \\
$f_1$ & RELU & RELU \\
$\gamma$ & $0.1$ & $0.1$ \\
$\lambda$ & $0.01$ & $-$ \\
\hline
\end{tabular}
\caption{Networks Hyperparameters.}
\label{tabellaHyper}
\end{table}
\subsection{Numerical Simulations}
\label{simulations}
\begin{figure}[h!]
\centering
\includegraphics[width=1\columnwidth]{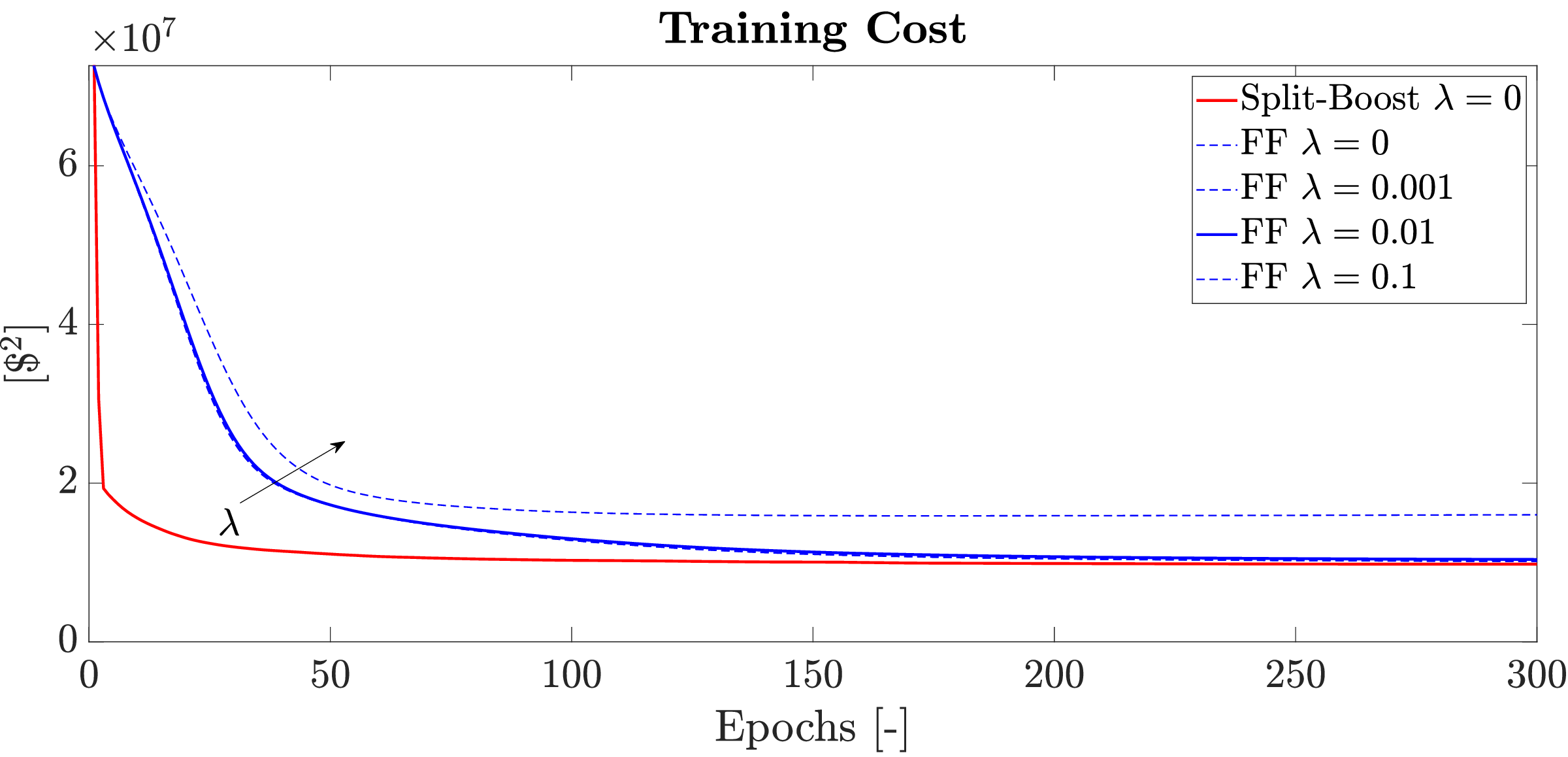}
\caption{Training cost: comparison between  \textit{split-boost} and  \textit{feed-forward} network for different values of the regularization hyper-parameter $\lambda$. Solid line is the best configuration of $\lambda$. }
\label{fig:training}
\end{figure}

In this section we show numerical insights about the training procedure of the split-boost network.
Figure \ref{fig:training} shows the trend of the training cost along the training epochs for the two networks. The split-boost network is represented by the solid red line, the feed-forward is represented by the blue lines for varying values of  $\lambda$. The solid blue line corresponds to the best $\lambda$ identified (see Table \ref{tabellaHyper}). As $\lambda$ increases, the training cost of the feed-forward network increases. For smaller values of $\lambda$, it decreases. Split-boost training cost converges to values close to those of the feed-forward regime for a substantially lower number of epochs, $E_{TB}^*=50$ epochs against $E_{FF}^*=200$ epochs. This allows us to conclude that the split-boost procedure, in this case, can converge to the maximum information content extractable from the training set, in a smaller number of epochs, and with an implicit regularization effect. 

Notice that the selection of the best number of epochs (implementing early stopping strategy) is performed looking at validation cost as discussed in Section \ref{details} for both the networks. After the best epoch retrieval, both the networks are re-trained considering as training set the union  of the previous training and validation sets. 

\begin{figure}[h!]
\centering
\includegraphics[width=.8\columnwidth]{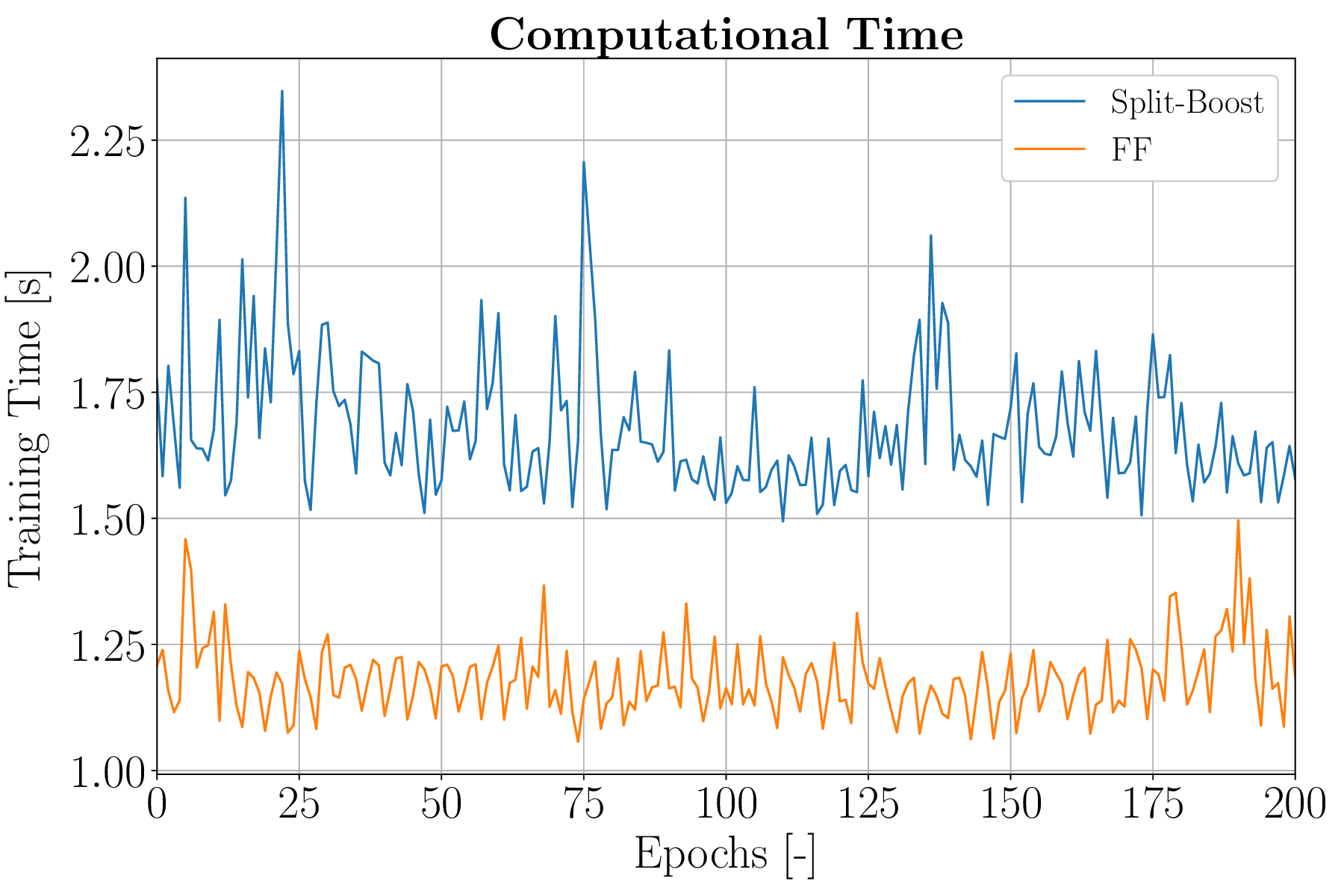}
\caption{Training time $[s]$ per epoch: comparison between  \textit{split-boost} and  \textit{feed-forward} network over 200 epochs. }
\label{fig:trainingTime}
\end{figure}
In Figure \ref{fig:trainingTime} the computational training time required by the 2 strategies is shown over 200 epochs. Split-boost strategy shows an average training time of $T_{SB} = 1.679 \, s$ while the feed-forward of $T_{FF} = 1.179 \, s$. On average, each epoch of the split-boost takes $42\%$ more time than the feed-forward. However, considering the training convergence shown in Figure \ref{fig:training}, which highlights that split-boost training cost converges at the regime in $E_{SB}^*=50$ epochs against the $E_{FF}^*=200$ of the feed-forward, leads to the average computational requirements of:
\begin{equation}
    E_{SB}^* \cdot T_{SB} = 83.95 \,s \leq  E_{FF}^* \cdot T_{FF} = 235.8 \,s. 
\end{equation}
%
\begin{figure*}[h!]
\centering
\includegraphics[width=0.9\textwidth]{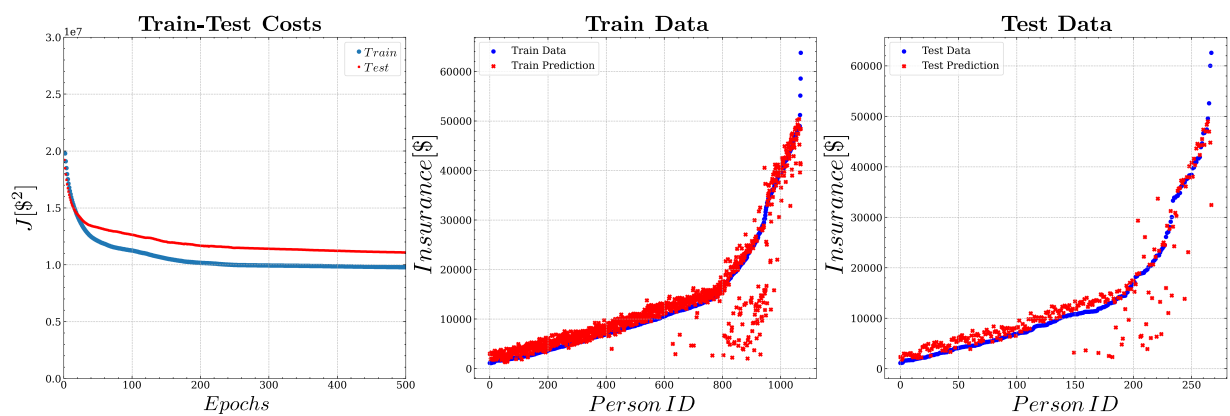}
\caption{Training and test cost for the split-boost (left panel);  predicted vs estimated output in the training dataset (middle panel);  predicted vs estimated output in the test dataset (right panel); }
\label{fig:regressionpb}
\end{figure*}
In Figure \ref{fig:regressionpb} the re-training procedure of the split-boost neural network is shown. On the left the training and test costs are shown. If compared with Figure \ref{fig:training}, the number of epochs to which the training cost regime is reached is higher. This depends on the fact that the training set is larger (it includes also the previous validation set). In the middle, the plot of the regression prediction versus target values for each person in the training set is shown. On the right plot, regression prediction versus target values for test data is shown. In the middle and right panels it is shown that the neural network can map successfully the non-linear relationship between the features collected in Table \ref{dataFeatures} and the regression target. There are some patients whose characteristics escape mapping: with similar features compared to the remaining patients, a higher medical insurance cost is attributed to them by the experts.

\begin{figure}[htbp!]
\centering
\includegraphics[width=.8\columnwidth]{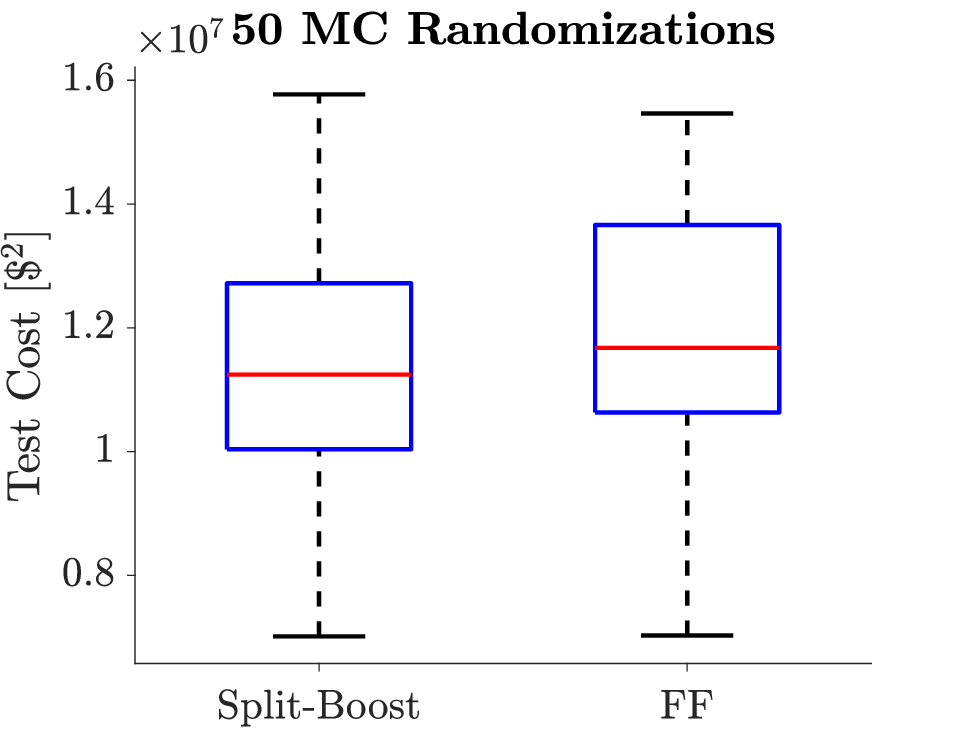}
\caption{Test cost (after re-training procedure)   for 50 randomizations of the extraction of training, validation and test sets.}
\label{fig:test_cost_regression}
\end{figure}
To evaluate the performance of the split-boost strategy with respect to the best regularized feed-forward neural network with $\lambda=0.01$, derived from the sensitivity analysis carried on the regularizing term for the traditional FFNN in Fig. \ref{fig:tuning}, 50 Monte Carlo randomizations of the dataset were performed, extracting 50 different combinations of training, validation and test sets. The results obtained on the test set were collected within the boxplots in Figure \ref{fig:test_cost_regression}. The test cost $J_{ts}= \frac{1}{2N_{ts}}(Y_v-\hat{Y}_{ts})^2$ obtained after the Monte Carlo randomizations in the case of the split-boost strategy is statistically lower. Split-boost test cost is lower in $72\%$ of the cases. This proves that, with statistical evidence, the split-boost neural network overcomes the feed-forward neural network also in terms of prediction accuracy.

\section{CONCLUSIONS}
In this article, we have shown an alternative training approach for  feed-forward neural networks. We have called this strategy ``split-boost'' to recall the idea that dividing (\textit{split}) the dataset and combining the  subsets \textit{might} lead to an improvement (\textit{boost}) in performance. 

In the considered real-world case study, the ``split-boost''  approach turns out to:  lead to higher predictive performance than traditional training; and  be computationally advantageous since in the training phase it converges within a smaller number of epochs, although the computational time per epoch is greater. 
The proposed strategy also implicitly  counteracts overfitting.

Future activities will focus on an extensive validation of the proposed training strategy, as well as on its generalization and extension to multi-layer networks.

\bibliography{bibliografia} 

\begin{thebibliography}{30}
\providecommand{\natexlab}[1]{#1}
\providecommand{\url}[1]{\texttt{#1}}
\providecommand{\urlprefix}{URL }
\expandafter\ifx\csname urlstyle\endcsname\relax
  \providecommand{\doi}[1]{doi:\discretionary{}{}{}#1}\else
  \providecommand{\doi}{doi:\discretionary{}{}{}\begingroup
  \urlstyle{rm}\Url}\fi

\bibitem[{Arlot and Celisse(2010)}]{Arlot}
Arlot, S. and Celisse, A. (2010).
\newblock A survey of cross-validation procedures for model selection.

\bibitem[{Bartlett et~al.(2006)Bartlett, Jordan, and McAuliffe}]{Bartlett}
Bartlett, P., Jordan, M., and McAuliffe, J. (2006).
\newblock Convexity, classification, and risk bounds.
\newblock \emph{Journal of the American Statistical Association}, 101(473),
  138--156.

\bibitem[{Bishop(1995)}]{Bishop}
Bishop, C. (1995).
\newblock Neural networks for pattern recognition.
\newblock \emph{Oxford University Press}.

\bibitem[{Bottou(1991)}]{Bottou}
Bottou, L. (1991).
\newblock Stochastic gradient learning in neural networks.
\newblock \emph{Proceedings of Neuro-Nımes}, 91.8, 12.

\bibitem[{Fahlman and Lebiere(1989)}]{Fahlman}
Fahlman, S. and Lebiere, C. (1989).
\newblock The cascade-correlation learning architecture.
\newblock \emph{Advances in neural information processing systems}, 2.

\bibitem[{Geisser(1975)}]{Geisser}
Geisser, S. (1975).
\newblock The predictive sample reuse method with applications.
\newblock \emph{Journal of the American statistical Association}, 70.350,
  320--328.

\bibitem[{Glorot et~al.(2011)Glorot, Bordes, and Bengio}]{Glorot}
Glorot, X., Bordes, A., and Bengio, Y. (2011).
\newblock Deep sparse rectifier neural networks.
\newblock \emph{Proceedings of the fourteenth international conference on
  artificial intelligence and statistics, JMLR Workshop and Conference
  Proceedings}.

\bibitem[{Goodfellow et~al.(2016)Goodfellow, Bengio, and
  Courville}]{Goodfellow}
Goodfellow, I., Bengio, Y., and Courville, A. (2016).
\newblock \emph{Deep learning}.
\newblock MIT press.

\bibitem[{Gupta and et~al.(2015)}]{Gupta}
Gupta, S. and et~al. (2015).
\newblock Deep learning with limited numerical precision.
\newblock \emph{International conference on machine learning, PMLR}.

\bibitem[{Hecht-Nielsen(1987)}]{Hecht-Nielsen}
Hecht-Nielsen, R. (1987).
\newblock Kolmogorov’s mapping neural network existence theorem.

\bibitem[{Hinton et~al.(2015)Hinton, Vinyals, and Dean}]{Hinton}
Hinton, G., Vinyals, O., and Dean, J. (2015).
\newblock Distilling the knowledge in a neural network.
\newblock \emph{arXiv:1503.02531}.

\bibitem[{Huang and Guang-Bin(2015)}]{Huang}
Huang and Guang-Bin (2015).
\newblock What are extreme learning machines? filling the gap between frank
  rosenblatt’s dream and john von neumann’s puzzle.
\newblock \emph{Cognitive Computation}, (7), 263--278.

\bibitem[{Huang et~al.(2021)Huang, Xuyang, and et~al.}]{Huang2}
Huang, Xuyang, and et~al. (2021).
\newblock A backpropagation extreme learning machine approach to fast training
  neural network-based side-channel attack.
\newblock \emph{Asian Hardware Oriented Security and Trust Symposium
  (AsianHOST),IEEE}.

\bibitem[{Ioffe and Szegedy(2015)}]{Ioffe}
Ioffe, S. and Szegedy, C. (2015).
\newblock Batch normalization: Accelerating deep network training by reducing
  internal covariate shift.
\newblock \emph{International conference on machine learning, pmlr}.

\bibitem[{Keskar and et~al.(2016)}]{Keskar}
Keskar, N. and et~al. (2016).
\newblock On large-batch training for deep learning: Generalization gap and
  sharp minima.
\newblock \emph{arXiv:1609.04836}.

\bibitem[{Kohavi(1995)}]{Kohavi}
Kohavi, R. (1995).
\newblock A study of cross-validation and bootstrap for accuracy estimation and
  model selection.
\newblock \emph{Ijcai}, 14.2, 35--47.

\bibitem[{Krizhevsky et~al.(2017)Krizhevsky, Sutskever, and Hinton}]{Sutskever}
Krizhevsky, A., Sutskever, I., and Hinton, G. (2017).
\newblock Imagenet classification with deep convolutional neural networks.
\newblock \emph{Communications of the ACM}, 60(6), 84--90.

\bibitem[{Lantz(2013)}]{book}
Lantz, B. (2013).
\newblock Machine learning with r: Learn how to use r to apply powerful machine
  learning methods and gain an insight into real world applications.
\newblock \emph{Livery Place}.

\bibitem[{LeCun and et~al.(1998)}]{LeCun}
LeCun, Y. and et~al. (1998).
\newblock Gradient-based learning applied to document recognition.
\newblock \emph{Proceedings of the IEEE}, 86.11, 2278--2324.

\bibitem[{Loshchilov and Hutter(2017)}]{Loshchilov}
Loshchilov, I. and Hutter, F. (2017).
\newblock Decoupled weight decay regularization.
\newblock \emph{arXiv:1711.05101}.

\bibitem[{Nair and Hinton(2010)}]{Nair}
Nair, V. and Hinton, G. (2010).
\newblock Rectified linear units improve restricted boltzmann machines.
\newblock \emph{Proceedings of the 27th international conference on machine
  learning (ICML-10)}.

\bibitem[{Paszke and et~al.(2019)}]{pytorch}
Paszke, A. and et~al. (2019).
\newblock Pytorch: An imperative style, high-performance deep learning library.
\newblock \emph{Advances in neural information processing systems}, 32.

\bibitem[{Rosenblatt(1958)}]{Rosenblatt}
Rosenblatt, F. (1958).
\newblock The perceptron: a probabilistic model for information storage and
  organization in the brain.
\newblock \emph{Psychological review}, 65.6, 386.

\bibitem[{Rumelhart et~al.(1986)Rumelhart, Hinton, and Williams}]{Rumelhart}
Rumelhart, D.E., Hinton, G., and Williams, R. (1986).
\newblock Learning representations by back-propagating errors.
\newblock \emph{Nature}, 323.6088, 533--536.

\bibitem[{Srivastava and et~al.(2014)}]{Srivastava}
Srivastava, N. and et~al. (2014).
\newblock Dropout: a simple way to prevent neural networks from overfitting.
\newblock \emph{The journal of machine learning research}, 15(1), 1929--1958.

\bibitem[{Tieleman and Hinton(2009)}]{hinton2}
Tieleman, T. and Hinton, G. (2009).
\newblock Using fast weights to improve persistent contrastive divergence.
\newblock \emph{Proceedings of the 26th annual international conference on
  machine learning}.

\bibitem[{Varma and Simon.(2006)}]{Varma}
Varma, S. and Simon., R. (2006).
\newblock Bias in error estimation when using cross-validation for model
  selection.
\newblock \emph{BMC bioinformatics}, 7.1, 1--8.

\bibitem[{Wan and et~al.(2013)}]{Krizhevsky}
Wan, L. and et~al. (2013).
\newblock Regularization of neural networks using dropconnect.
\newblock \emph{International conference on machine learning, PMLR}.

\bibitem[{Werbos(1988)}]{Werbos}
Werbos, P. (1988).
\newblock Generalization of backpropagation with application to a recurrent gas
  market model.
\newblock \emph{Neural networks}, 1.4, 339--356.

\bibitem[{Zhang and et~al.(2021)}]{Bengio}
Zhang, C. and et~al. (2021).
\newblock Understanding deep learning (still) requires rethinking
  generalization.
\newblock \emph{Communications of the ACM}, 64(3), 107--115.

\end{thebibliography}
\addtolength{\textheight}{-12cm}   




\end{document}